\documentclass{article} %
\usepackage{collas2023_conference,times}
\usepackage{easyReview}

\usepackage{amsmath,amsfonts,bm}

\def\eqref#1{equation~\ref{#1}}

\def\1{\bm{1}}

\DeclareMathAlphabet{\mathsfit}{\encodingdefault}{\sfdefault}{m}{sl}
\SetMathAlphabet{\mathsfit}{bold}{\encodingdefault}{\sfdefault}{bx}{n}

\DeclareMathOperator*{\argmax}{arg\,max}

\usepackage{hyperref}
\hypersetup{
    colorlinks=true,
    linkcolor=red,
    filecolor=magenta,
    urlcolor=blue,
    citecolor=purple,
    pdftitle={Improved OOD detection in SHELS},
    pdfpagemode=FullScreen,
    }

\usepackage{algorithm}
\usepackage{algpseudocode}
\usepackage{booktabs}
\usepackage{multirow}

\usepackage{natbib}

\usepackage{subcaption}

\usepackage{amssymb}
\usepackage{enumitem}
\usepackage{setspace}
\usepackage{todonotes}
\newcounter{mycomment}

\title{Continual Improvement of \\Threshold-Based Novelty Detection}

\author{Abe Ejilemele \\
Computer Science and Artificial Intelligence Laboratory\\
Massachusetts Institute of Technology\\
\texttt{aoejile@mit.edu} 
\And %
Jorge Mendez-Mendez \\
Computer Science and Artificial Intelligence Laboratory\\
Massachusetts Institute of Technology\\
\texttt{jmendez@csail.mit.edu} 
}

\collasfinalcopy %

\begin{document}

\maketitle

\begin{abstract}
When evaluated in dynamic, open-world situations, neural networks struggle to detect unseen classes. This issue complicates the deployment of continual learners in realistic environments where agents are not explicitly informed when novel categories are encountered. A common family of techniques for detecting novelty relies on thresholds of similarity between observed data points and the data used for training. However, these methods often require manually specifying (ahead of time) the value of these thresholds, and are therefore incapable of adapting to the nature of the data. We propose a new method for automatically selecting these thresholds utilizing a linear search and leave-one-out cross-validation on the ID classes. We demonstrate that this novel method for selecting thresholds results in improved total accuracy on MNIST, Fashion MNIST, and CIFAR-10.
\end{abstract}

\section{Introduction}

Humans and other natural learners learn in a continual fashion. By accumulating a knowledge base of learned concept representations, they are able to recognize when they see something novel. As they encounter new concepts, they incorporate these concepts' representations into their growing knowledge base. When incorporating these ideas into machine learning methods, one key challenge is automatically determining when a new concept is encountered in order to learn and incorporate its representation. We propose a mechanism for this type of OOD detection that treats multiple ID classes as OOD in turn to extract the best possible OOD detection threshold for \textit{future} classes.

Our method operates in the continual learning setting, where models are trained on a sequence of tasks (in our case, classes) consecutively, and must maintain performance on previous classes without forgetting. Most methods in this space assume that an oracle informs the agent of the identity of the task, and such methods typically use a set of task-specific parameters~\citep{kirkpatrick2017overcoming,zenke2017continual,li2017learning,nguyen2018variational,serra2018overcoming,lopez2017gradient}. We focus instead in a setting where the agent encounters new concepts and must autonomously decide when novelty is encountered to trigger the accommodation of the new concepts, similar to the settings of \citet{ren2020wandering} and \citet{gummadi2022shels}. Other approaches have focused on OOD detection, but not on the accommodation of novelty into the model~\citep{hendrycks2016baseline,liang2017enhancing}. OOD detection can also be viewed as a form of change-point detection~\citep{Aminikhanghahi2017survey}.

We apply our technique to the sparse, high-level exclusive, low-level shared (SHELS) feature representation of \citet{gummadi2022shels}. The key idea behind SHELS is to represent classes as exclusive feature sets by using cosine normalization, while maintaining only the essential features of each representation at lower levels via group sparsity regularization. The exclusivity of high-level features enables detecting novelty as unseen sets of features, and the sparsity of lower-level features permits accommodating novelty by making use of spare network capacity. As many other novelty detection approaches, SHELS relies on a pre-defined threshold for categorizing a new data point as OOD. 

The main contributions of this work include:
\begin{enumerate}%
    \item We show empirically that there is a significant performance gap between the threshold pre-defined by SHELS and the optimal threshold obtained by looking ahead at the new OOD class and conducting a linear search on potential values of the threshold to maximize the performance on ID classes and the new OOD class simultaneously---note that this method ``cheats'' by observing OOD instances.
    \item We propose a mechanism for closing this performance gap by conducting a similar linear search, but instead using only ID data to compute the optimal threshold. Our proposed solution can be viewed as a continual hyper-parameter optimization method via leave-one-class-out cross-validation. 
    \item Our experiments demonstrate that our approach yields a good estimate of the optimal threshold found through ``cheating". We evaluate this new dynamic threshold search method in a within-dataset detection setting, where novelty detection is tricky since OOD and ID classes share similar features. 
\end{enumerate} 

\section{Background on SHELS}

\citet{gummadi2022shels} consider a setting in which each data point $x \in \mathbb{R}^d$ is represented by a set of features $f = g(x) \in \mathbb{R}^k$, which is the set of activations of the final hidden layer of a neural network. In this setting, a pair of exclusive sets of features is defined as a pair $f_1, f_2$ that satisfies the conditions $\underline{f_1} \setminus \underline{f_2} \neq \emptyset \wedge \underline{f_2} \setminus \underline{f_1} \neq \emptyset$, where $\setminus$ denotes the set difference operation and we have used $\underline{\cdot}$ to denote the treatment of a set of real values as a set via some thresholding operation that converts real numbers to binary membership values. \citet{gummadi2022shels} show that orthogonality of feature vectors is a stronger requirement than set exclusivity. In consequence, SHELS uses cosine normalization to space ID classes along the surface of a $k$-dimensional unit ball, encouraging small cosine similarity (and therefore orthogonality and exclusivity) between ID classes. Intuitively, this should encourage OOD classes to also be represented as exclusive feature sets. 

Given a trained (exclusive) feature representation, SHELS determines whether a new input is OOD via a threshold on the class similarity score to each class $c\in\{1,\ldots,C\}$:
\begin{equation}
s_c = W_{[:, c]}^{L^\top}f^{L -1}(x) \label{eq:1}\enspace,
\end{equation} 
where $W_[:, c]^{L}$ are the weights of the output layer of the network for class $c$. 
Inputs whose features lie in the same direction as a class's weights will have a high score, and those whose features lie in a different direction will have low scores. A sample is considered OOD if its similarity score is below a threshold $th_c = \mu_c - \sigma_c$ for all ID classes $c$, where $\mu_c$ and $\sigma_c$ are the mean and standard deviation of the similarity scores over the training data for class $c$. 

Once an OOD instance is detected, SHELS incorporates the novel concept into its knowledge base. In order to do this without forgetting the ID classes (i.e., without catastrophic forgetting, \citealp{mccloskey1989catastrophic}), SHELS imposes two forms of regularization. First, a group sparsity penalty encourages sparse representations at the low levels of the network, leaving sufficient capacity for subsequent classes to be learned. Second, weights that were previously used for ID classes are (soft) frozen by penalizing deviations from their previous values. 

\section{Approach}

We propose a method to replace fixed, previously chosen thresholds for continual OOD detection with dynamically chosen ones based on the nature of the data. We apply our technique to threshold selection in SHELS in a natural continual learning setting where new classes are encountered incrementally and the learner must automatically determine when that occurs before accommodating the new concepts.

Intuitively, a good threshold for OOD detection should be sufficiently low to correctly classify ID data, but still large enough to correctly detect OOD points. Our approach seeks to find this middle ground. We also emphasize that a good method for threshold selection should be computationally efficient, since it must be executed repeatedly as additional data is presented to the learner. %

\subsection{Cheating to find an optimal threshold}
\label{sec:cheatingMethod}

To motivate our approach, we begin by describing how an optimal threshold can be obtained if the learner knows ahead of time what the OOD data will look like. In addition to serving as a thought exercise to guide our mechanism design, this cheating variant of our method will also serve to demonstrate that there is much room for improvement compared to the fixed threshold defined by SHELS. 

We assume that there is a scoring function $s_c(x): \mathbb{R}^k \mapsto \mathbb{R}$ that measures how close an input $x$ is to being considered part of class $c$ (e.g., the SHELS scoring function in Equation~\ref{eq:1}). Given a model trained on a set of ID classes, we first compute $s_c(x)$ for all $x$ in the training data for class $c$, for all classes $c$. We do this only over \textit{correctly classified} ID points to avoid conflating ID prediction ability with OOD detection. If we look ahead at a sample of OOD data, we can compute the similarity scores of each data point to each class $c$. We can then simply perform a linear search over values of the threshold over the computed similarity scores (of both ID and OOD data) and select the threshold value that maximizes the performance metric. %

\subsection{Our continual OOD threshold selection method}

Our mechanism for choosing an OOD threshold is motivated by the cheating approach of Section~\ref{sec:cheatingMethod}, but uses only the ID classes---without looking ahead at the OOD classes---to find an approximation of the optimal threshold. The process is divided into two stages. In the first stage, we treat each ID class as OOD in turn in a leave-one-class-out cross-validation fashion. For a given left-out OOD class, we train a model on the remaining ID classes and compute the optimal threshold for the left-out class following the cheating mechanism. We then average the thresholds obtained for each left-out class to compute the threshold for the first (true) OOD class. 

Thereafter, the model continually observes data and uses its OOD detection threshold to determine whether the data is ID or OOD. Following the evaluation protocol of \citet{gummadi2022shels}, whenever a batch of data is determined to contain OOD instances, the agent requests a training set of the OOD class. At this point---and, critically, before accommodating the OOD class into the model and making it now-ID---the agent uses the observed OOD data to find what the optimal threshold \textit{would have been} for this new OOD class. We then update our previous threshold by averaging it with the new one, to use for the next round of OOD detection, and accommodate the current OOD class.

\subsection{Application of dynamic threshold selection to SHELS}

The original threshold used in SHELS is given by $th_c^{\mathrm{one}} = \mu_c - \sigma_c$. We perform our search in terms of numbers of standard deviations from the mean similarity score: $th_c = \mu_c - \eta \sigma_c$, where $\eta$ is the hyperaparameter that our method will search for. Our goal is to find $\eta^*$ that yields the optimal threshold $th_c^*$.

SHELS correctly classifies a datapoint from class $c$ if:
\begin{equation*}
    \argmax_c s_c(x) = c \quad\quad \text{and} \quad\quad \max_c s_c(x) > th_c\enspace.
\end{equation*}
Similariy, SHELS correctly detects an OOD point if:
\begin{equation*}
    \argmax_c s_c(x) = c \quad\quad \text{and} \quad\quad \max_c s_c(x) < th_c\enspace.
\end{equation*}

To conduct our search over values of $\eta$, it will be useful to define $Z_c(x)^\prime=-Z_c(x)=\frac{\mu_c - s_c(x)}{\sigma_c}$ as the negative Z-score of the similarity scores of class $c$ over correctly classified training points. The search space for possible values of $\eta$ consists of the $Z_c(x)^\prime$ scores of all correctly classified ID points of class $c$ and all OOD points. Then, an ID point is correctly classified if $Z_c(x)^\prime < \eta$, and an OOD point is detected correctly if $Z_c(x)^\prime > \eta$, for any chosen $\eta$. 

In our experiments, we use two performance metrics to select the optimal threshold during search: total accuracy, defined as $\frac{\mathrm{ID\ accuracy} + \mathrm{OOD\ accuracy}}{2}$~\citep{gummadi2022shels} and the G-mean metric, defined as $\sqrt{\mathrm{ID\ accuracy} \cdot \mathrm{OOD\ accuracy}}$~\citep{Aminikhanghahi2017survey}. We do this because often there is a substantial imbalance between ID and OOD data, making G-mean a more robust performance measure. 

In SHELS, it is likely that the accommodation of new OOD classes modifies model parameters in a way that affects the previous ID classes' activations. For this reason, we recompute the thresholds $th_c$ for all previous $c$'s by obtaining the updated activations for the training data of those previous classes to recompute the similarity scores $s_c$. Note that, while this process does require storing all previous training data, it does not require any training on previous data or recomputing of the hyperparameter $\eta$. 

Critically, note that our process never requires training more than a single model continually after the initial cross-validation stage, making it efficient in the number of classes encountered continually. This is due to the fact that the search over possible threshold values does not require training a separate model for each threshold choice, but is instead applied post hoc on trained models. Our method could also be used to select other post hoc hyperparameters efficiently in a continual OOD detection and accommodation setting.

\begin{figure}[t!]
\centering
    \begin{subfigure}[b]{\linewidth}
        \centering
        \includegraphics[height=1.2cm, trim={3cm 8.8cm 1.7cm 1.6cm}, clip]{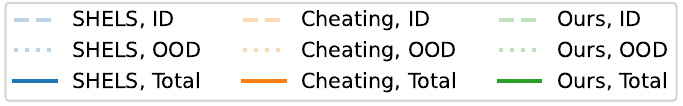}
    \end{subfigure}\\    \begin{subfigure}[b]{0.33\linewidth}
        \centering
        \includegraphics[height=3.7cm, trim={0.8cm 0.8cm 0.8cm 0.8cm}, clip]{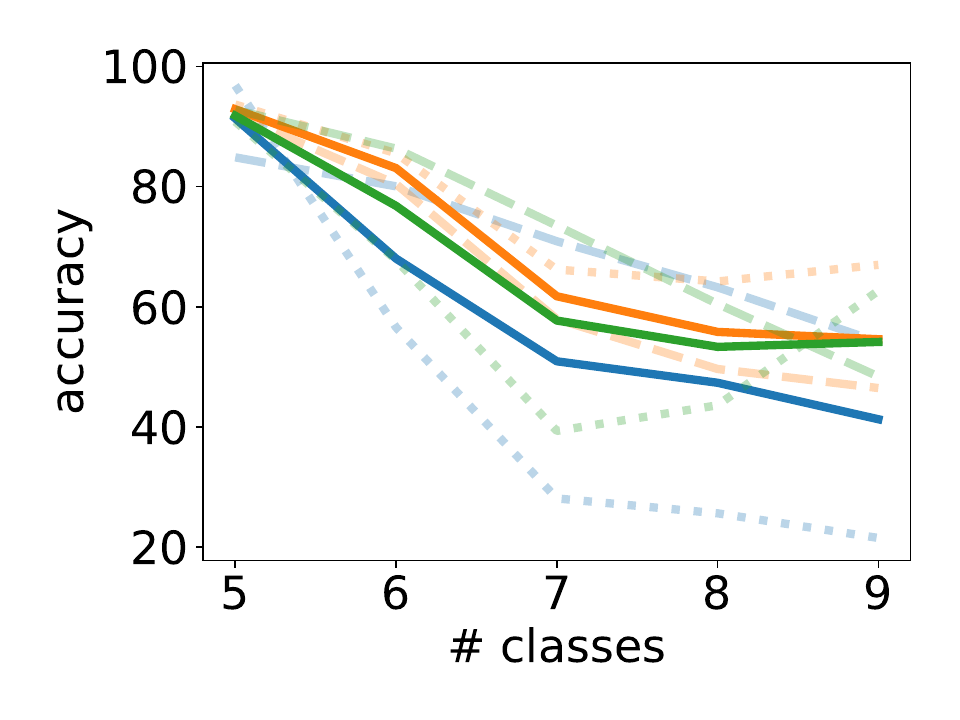}
        \caption{MNIST}
        \label{fig:mnist}
    \end{subfigure}%
    \begin{subfigure}[b]{0.33\linewidth}
        \centering
        \includegraphics[height=3.7cm, trim={1.6cm 0.8cm 0.8cm 0.8cm}, clip]{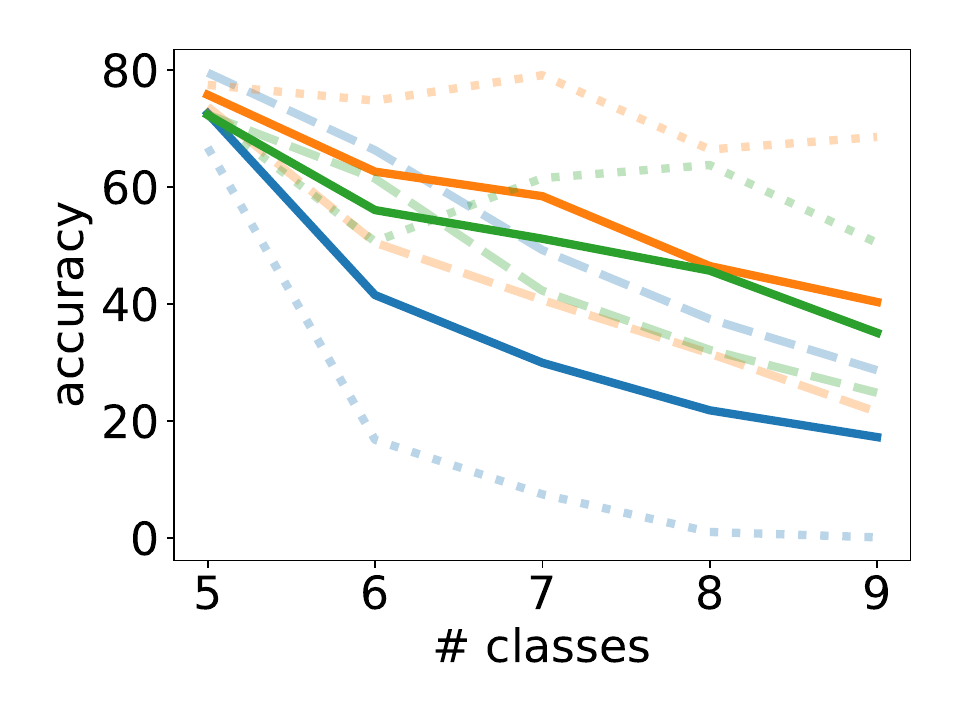}
        \caption{Fashion MNIST}
        \label{fig:fmnist}
    \end{subfigure}%
    \begin{subfigure}[b]{0.33\linewidth}
        \centering
        \includegraphics[height=3.7cm, trim={1.6cm 0.8cm 0.8cm 0.8cm}, clip]{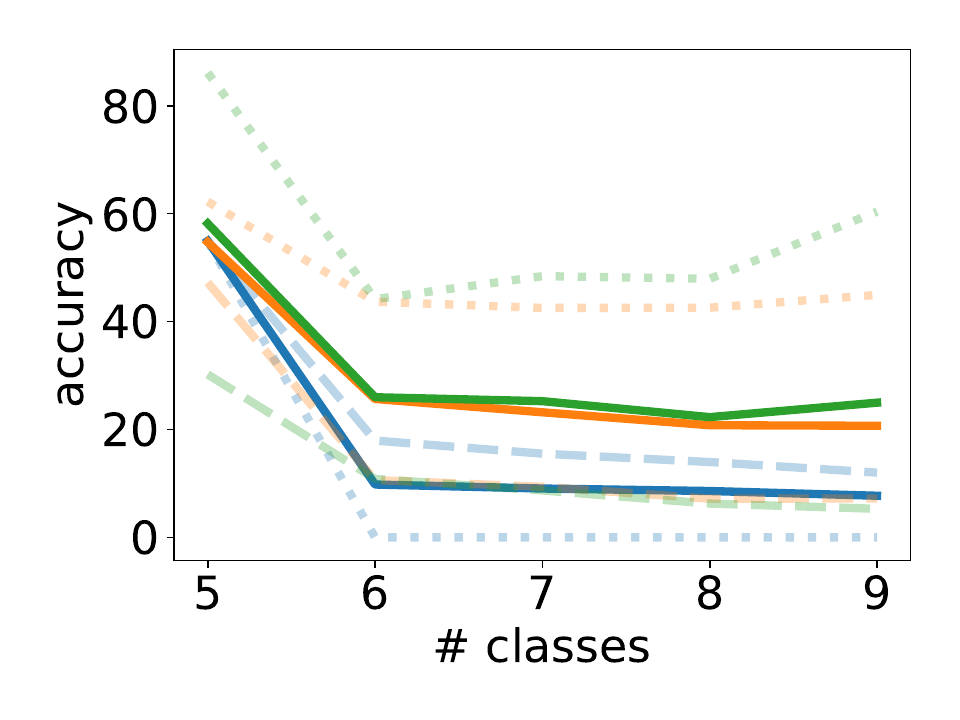}
        \caption{CIFAR-10}
        \label{fig:cifar}
    \end{subfigure}
    \caption{Total accuracy, ID accuracy, and OOD accuracy for SHELS, the cheating baseline, and our method for dynamic threshold selection. Both the cheating method and our approach consistently outperform SHELS in total accuracy and OOD accuracy, albeit at the cost of (slightly) reduced ID accuracy. Averaged over 10 random seeds.}
    \label{fig:results}
\end{figure}

\section{Empirical evaluation}

Our experiments were carried out to show that 1) the threshold obtained by looking ahead at the OOD class (cheating) substantially outperforms the threshold used by SHELS, and 2) the threshold obtained by our method is competitive with the cheating threshold. 

\subsection{Experimental setting}

\paragraph{Datsets}  The algorithms were evaluated using MNIST \citep{lecun1998gradient}, Fashion MNIST \citep{xiao2017fashion}, and CIFAR-10 \citep{krizhevsky2009learning}. MNIST,
FMNIST, and CIFAR10 each consist of 10 classes.

\paragraph{Training details}  Model architectures and hyperparameters matched those used by \citet{gummadi2022shels}. We trained MNIST for 10 epochs, Fashion MNIST for 20 epochs, and CIFAR-10 for 35 epochs, using a batch size of 32.  

\paragraph{Evaluation protocol} Our experiments evaluated the model's ability to both detect OOD data with high accuracy and continue to classify ID data. We followed the protocol of \citet{gummadi2022shels}, which trains the model on ID data, and then feeds the model one new OOD class at a time, which is automatically detected and subsequently accommodated into the model. We focused on within-dataset novelty detection---i.e., detecting novelty among classes with high feature similarity. For each dataset, we randomly sampled $C=5$ classes as ID and treated the rest as OOD. Each experiment was repeated 10 times with varying random seeds controlling the choice of ID classes and model initialization. We used all training and validation data for training and evaluated ID accuracy on the test data and OOD accuracy on the training data---since the OOD training data is unseen by the model, it is a fair evaluation set.

\subsection{Results}

Figure~\ref{fig:results} summarizes our main results. Clearly, both the cheating and non-cheating methods achieve substantially higher total accuracy than the original SHELS approach. These results were obtained by optimizing for G-mean during the threshold selection stage. Notably, the non-cheating method closely matched the performance of the method that cheats. It is also apparent that our method's threshold leads to improved OOD accuracy in all cases at the cost of slightly reduced ID accuracy---their balance, as computed via the total accuracy, is consistently better. 

Complete results in Appendix~\ref{section: B},  that, for MNIST, optimizing for the G-mean during search leads to better performance than total accuracy, as expected due to the imbalance between ID and OOD instances. Moreover, statistical significance tests in Appendix~\ref{section: B} demonstrate that our method's improvements are in many cases significant.

\section{Conclusions}
We proposed a method for automatically selecting the threshold for OOD detection in a continual setting, where novel classes are encountered consecutively and incorporated into the model. Our mechanism searches for thresholds that would perform well if the ID classes were OOD, and then extends these estimates to the next OOD class. Our experiments demonstrate consistent improvements in total accuracy over the fixed threshold chosen by SHELS. Two downsides of our approach are the multiple training rounds required during the (initial) cross-validation stage, and the requirement that all ID data is stored to recompute the activations for future thresholds. Improving our method's efficiency in either of these dimensions would yield an approach that is better suited for continual deployment.

\bibliography{References}
\bibliographystyle{collas2023_conference}

\newpage
\appendix

\section{Experimental results}\label{section: B}

Tables~\ref{table 1}--\ref{table 4} display the full set of results obtained for our experiments on MNIST, Fashion MNIST, and CIFAR-10, including statistical testing. 

\begin{table}[H]
\centering
\caption{Avg ± std. dev. accuracy results for MNIST optimizing for total accuracy across 10 random seeds, and Student's t-test significance results against the 1 std dev baseline with Holm-Bonferroni corrections. Statistically significant results after Holm-Bonferroni corrections are denoted for $p < .05$ as *, $p < .01$ as ** and $p < .001$ as ***.}
\label{table 1}
\begin{tabular}{ll|lll|llll|ll}
\toprule
$\#$ ID & Metric &     SHELS &         Cheating &   Ours & \multicolumn{2}{c}{$p_{\mathrm{cheat}}$} & \multicolumn{2}{c|}{$p_{\mathrm{ours}}$} &  $\eta_{\mathrm{cheat}}$ &  $\eta_{\mathrm{ours}}$ \\
classes & & & & &  unadj. & adj. & unadj. & adj. \\
\midrule
\midrule
\multirow{3}{*}{5} & ID &   84.91 {\tiny± 0.46} &  96.61 {\tiny± 1.11}*** &  96.67 {\tiny± 0.8}*** &            0.00 &                      0.00 &                0.00 &                0.00 &      \multirow{3}{*}{2.11} &               \multirow{3}{*}{2.09} \\
 & OOD &   97.43 {\tiny± 2.38} &   83.11 {\tiny± 7.38}** &  80.52 {\tiny± 8.44}** &            0.00 &                      0.00 &                0.00 &                0.00 &    & \\
 & Total &    91.7 {\tiny± 1.35} &     89.38 {\tiny± 4.16} &     88.05 {\tiny± 4.4} &            0.93 &                      1.00 &                0.98 &                1.00 &    & \\
\midrule
\multirow{3}{*}{6} & ID &   80.95 {\tiny± 3.47} &   88.87 {\tiny± 3.95}** &  90.5 {\tiny± 3.42}*** &            0.00 &                      0.00 &                0.00 &                0.00 &      \multirow{3}{*}{1.89} &               \multirow{3}{*}{2.10} \\
 & OOD &  50.88 {\tiny± 24.01} &    60.11 {\tiny± 23.64} &   46.42 {\tiny± 25.94} &            0.42 &                      1.00 &                0.71 &                1.00 & & \\
 & Total &  65.52 {\tiny± 12.11} &    74.33 {\tiny± 11.55} &    68.0 {\tiny± 13.34} &            0.07 &                      0.53 &                0.34 &                1.00 & & \\
\midrule
\multirow{3}{*}{7} & ID &   72.67 {\tiny± 3.08} &      77.9 {\tiny± 3.5}* &  78.76 {\tiny± 2.25}** &            0.00 &                      0.04 &                0.00 &                0.00 &      \multirow{3}{*}{2.17} &               \multirow{3}{*}{1.99} \\
 & OOD &  25.63 {\tiny± 23.94} &    43.92 {\tiny± 23.86} &   22.24 {\tiny± 18.26} &            0.12 &                      0.73 &                0.74 &                1.00 & & \\
 & Total &   50.59 {\tiny± 11.1} &      62.3 {\tiny± 9.55} &    52.33 {\tiny± 8.88} &            0.01 &                      0.13 &                0.36 &                1.00 & & \\
\midrule
\multirow{3}{*}{8}  & ID &    64.32 {\tiny± 3.5} &     69.34 {\tiny± 3.56} &    69.26 {\tiny± 3.54} &            0.01 &                      0.08 &                0.01 &                0.08 &      \multirow{3}{*}{2.57} &               \multirow{3}{*}{2.08} \\
 & OOD &  25.77 {\tiny± 22.61} &    29.65 {\tiny± 22.52} &   25.01 {\tiny± 19.39} &            0.72 &                      1.00 &                0.94 &                1.00 & & \\
 & Total &   48.06 {\tiny± 9.03} &    52.63 {\tiny± 10.25} &    50.53 {\tiny± 9.51} &            0.16 &                      0.82 &                0.29 &                1.00 & & \\
\midrule
\multirow{3}{*}{9} & ID &   57.72 {\tiny± 3.25} &     62.33 {\tiny± 3.38} &    62.39 {\tiny± 3.29} &            0.01 &                      0.09 &                0.01 &                0.08 &      \multirow{3}{*}{2.59} &               \multirow{3}{*}{2.33} \\
 & OOD &    22.7 {\tiny± 17.9} &    31.68 {\tiny± 28.97} &   26.01 {\tiny± 28.56} &            0.44 &                      1.00 &                0.77 &                1.00 & & \\
 & Total &   43.92 {\tiny± 7.83} &    50.29 {\tiny± 11.43} &   48.03 {\tiny± 11.76} &            0.09 &                      0.65 &                0.20 &                1.00 & & \\
\bottomrule
\end{tabular}
\end{table}

\begin{table} [H]
\centering
\caption{Avg ± std. dev. accuracy results for MNIST optimizing for G-mean across 10 random seeds. Student's t-test significance results against the 1 std dev baseline with Holm-Bonferroni corrections. Statistically significant results after Holm-Bonferroni corrections are denoted for $p < .05$ as *, $p < .01$ as ** and $p < .001$ as ***.}
\label{table 2}
\begin{tabular}{ll|lll|llll|ll}
\toprule
$\#$ ID & Metric &     SHELS &         Cheating &   Ours & \multicolumn{2}{c}{$p_{\mathrm{cheat}}$} & \multicolumn{2}{c|}{$p_{\mathrm{ours}}$} &  $\eta_{\mathrm{cheat}}$ &  $\eta_{\mathrm{ours}}$ \\
classes & & & & &  unadj. & adj. & unadj. & adj. \\
\midrule
\midrule
\multirow{3}{*}{5} & ID    &   84.86 {\tiny± 0.42} &   91.97 {\tiny± 3.25}*** &  92.76 {\tiny± 1.95}*** & 0.00 &                      0.00 &                0.00 &                          0.00 &         \multirow{3}{*}{1.55} &               \multirow{3}{*}{1.56} \\
& OOD &   96.82 {\tiny± 2.94} &      93.64 {\tiny± 2.86} &     90.81 {\tiny± 6.47} & 0.03 &                      0.10 &                0.02 &                          0.14 & & \\
& Total &   91.35 {\tiny± 1.64} &     92.86 {\tiny± 2.87}* &     91.78 {\tiny± 3.21} & 0.00 &                      0.02 &                0.23 &                          0.90 & & \\
\multirow{3}{*}{6} & ID    &   80.04 {\tiny± 3.81} &      80.42 {\tiny± 6.14} &      86.32 {\tiny± 4.5} & 0.88 &                      0.88 &                0.01 &                          0.06 &         \multirow{3}{*}{1.17} &               \multirow{3}{*}{1.56} \\
& OOD &   56.58 {\tiny± 27.8} &     85.54 {\tiny± 12.8}* &    67.79 {\tiny± 29.03} & 0.01 &                      0.04 &                0.41 &                          0.90 & & \\
& Total &  68.02 {\tiny± 13.71} &     83.07 {\tiny± 8.18}* &    76.79 {\tiny± 14.67} & 0.00 &                      0.02 &                0.01 &                          0.12 & & \\
\multirow{3}{*}{7} & ID    &     70.9 {\tiny± 4.1} &    57.95 {\tiny± 10.38}* &      73.5 {\tiny± 3.71} & 0.00 &                      0.02 &                0.17 &                          0.88 &         \multirow{3}{*}{0.59} &               \multirow{3}{*}{1.36} \\
& OOD &   28.12 {\tiny± 23.4} &   66.19 {\tiny± 13.78}** &    39.35 {\tiny± 27.38} & 0.00 &                      0.01 &                0.36 &                          0.90 & & \\
& Total &   50.93 {\tiny± 9.34} &    61.75 {\tiny± 10.37}* &    57.71 {\tiny± 11.95} & 0.00 &                      0.02 &                0.02 &                          0.14 & & \\
\multirow{3}{*}{8} & ID   &   63.24 {\tiny± 4.19} &   49.65 {\tiny± 3.92}*** &     60.48 {\tiny± 5.05} & 0.00 &                      0.00 &                0.22 &                          0.90 &         \multirow{3}{*}{0.40} &               \multirow{3}{*}{0.98} \\
& OOD &  25.64 {\tiny± 16.43} &  64.21 {\tiny± 12.53}*** &    43.61 {\tiny± 11.84} & 0.00 &                      0.00 &                0.02 &                          0.14 & & \\
& Total &   47.36 {\tiny± 6.59} &     55.82 {\tiny± 6.64}* &      53.33 {\tiny± 5.4} & 0.00 &                      0.01 &                0.01 &                          0.10 & & \\
\multirow{3}{*}{9} & ID   &    54.15 {\tiny± 6.4} &     46.48 {\tiny± 10.25} &      48.4 {\tiny± 6.84} & 0.07 &                      0.15 &                0.08 &                          0.49 &         \multirow{3}{*}{0.56} &               \multirow{3}{*}{0.69} \\
& OOD &  21.54 {\tiny± 23.55} &   66.98 {\tiny± 19.53}** &   62.67 {\tiny± 26.03}* & 0.00 &                      0.00 &                0.00 &                          0.03 & & \\
& Total &  41.29 {\tiny± 10.62} &     54.6 {\tiny± 13.75}* &   54.15 {\tiny± 12.66}* & 0.00 &                      0.02 &                0.00 &                          0.03 & & \\
\bottomrule
\end{tabular}
\end{table}

\begin{table} [H]
\centering
\caption{Avg ± std. dev. accuracy results for FMNIST optimizing for total accuracy across 10 random seeds, and Student's t-test significance results against the 1 std dev baseline with Holm-Bonferroni corrections. Statistically significant results after Holm-Bonferroni corrections are denoted for $p < .05$ as *, $p < .01$ as ** and $p < .001$ as ***.}
\label{table 6}
\begin{tabular}{ll|lll|llll|ll}
\toprule
$\#$ ID & Metric &     SHELS &         Cheating &   Ours & \multicolumn{2}{c}{$p_{\mathrm{cheat}}$} & \multicolumn{2}{c|}{$p_{\mathrm{ours}}$} &  $\eta_{\mathrm{cheat}}$ &  $\eta_{\mathrm{ours}}$ \\
classes & & & & &  unadj. & adj. & unadj. & adj. \\
\midrule
\midrule
\multirow{3}{*}{5} & ID &   79.52 {\tiny± 3.04} &     73.71 {\tiny± 10.42} &      72.38 {\tiny± 7.27} &      0.13 &        0.63 &    0.01 &      0.13  &         \multirow{3}{*}{0.82} &               \multirow{3}{*}{0.72} \\
& OOD &  66.75 {\tiny± 22.21} &     77.44 {\tiny± 10.32} &     72.23 {\tiny± 22.84} &      0.21 &        0.83 &    0.61 &      1.00  & & \\
& Total &  72.56 {\tiny± 11.32} &     75.74 {\tiny± 9.74}* &       72.3 {\tiny± 9.61} &      0.00 &        0.03 &    0.60 &      1.00  & & \\
\multirow{3}{*}{6} & ID &  66.27 {\tiny± 12.72} &     50.47 {\tiny± 12.48} &     61.42 {\tiny± 13.24} &      0.02 &        0.10 &    0.44 &      1.00  &         \multirow{3}{*}{0.68} &               \multirow{3}{*}{0.77} \\
& OOD &  16.76 {\tiny± 23.84} &  74.75 {\tiny± 15.26}*** &     50.65 {\tiny± 34.51} &      0.00 &        0.00 &    0.03 &      0.18  & & \\
& Total &  41.51 {\tiny± 15.44} &    62.61 {\tiny± 11.31}* &      56.04 {\tiny± 15.6} &      0.00 &        0.02 &    0.02 &      0.14  & & \\
\multirow{3}{*}{7} & ID &  49.21 {\tiny± 17.09} &     40.66 {\tiny± 18.45} &     42.26 {\tiny± 14.39} &      0.32 &        0.95 &    0.36 &      1.00  &         \multirow{3}{*}{0.81} &               \multirow{3}{*}{0.73} \\
& OOD &   7.45 {\tiny± 14.56} &   79.1 {\tiny± 17.38}*** &    61.5 {\tiny± 36.17}** &      0.00 &        0.00 &    0.00 &      0.01  & & \\
& Total &  29.93 {\tiny± 13.58} &   58.41 {\tiny± 14.92}** &    51.14 {\tiny± 18.88}* &      0.00 &        0.00 &    0.00 &      0.01  & & \\
\multirow{3}{*}{8} & ID &  37.42 {\tiny± 13.81} &     31.53 {\tiny± 11.57} &     32.14 {\tiny± 12.29} &      0.34 &        0.95 &    0.40 &      1.00  &         \multirow{3}{*}{0.98} &               \multirow{3}{*}{0.77} \\
& OOD &    1.01 {\tiny± 2.38} &   66.4 {\tiny± 25.53}*** &  63.74 {\tiny± 34.93}*** &      0.00 &        0.00 &    0.00 &      0.00  & & \\
& Total &   21.81 {\tiny± 8.61} &   46.47 {\tiny± 13.84}** &   45.68 {\tiny± 14.11}** &      0.00 &        0.00 &    0.00 &      0.00  & & \\
\multirow{3}{*}{9} & ID &  28.66 {\tiny± 16.73} &     21.52 {\tiny± 12.41} &     24.81 {\tiny± 12.82} &      0.32 &        0.95 &    0.59 &      1.00  &         \multirow{3}{*}{0.57} &               \multirow{3}{*}{0.87} \\
& OOD &    0.09 {\tiny± 0.21} &  68.54 {\tiny± 19.75}*** &   50.44 {\tiny± 35.52}** &      0.00 &        0.00 &    0.00 &      0.01  & & \\
& Total &   17.23 {\tiny± 10.1} &   40.33 {\tiny± 12.66}** &    35.06 {\tiny± 17.93}* &      0.00 &        0.00 &    0.00 &      0.03  & & \\
\bottomrule
\end{tabular}
\end{table}

\begin{table} [H]
\centering
\caption{Avg ± std. dev. accuracy results for CIFAR10 optimizing for G-mean across 10 random seeds, and Student's t-test significance results against the 1 std dev baseline with Holm-Bonferroni corrections. Statistically significant results after Holm-Bonferroni corrections are denoted for $p < .05$ as *, $p < .01$ as ** and $p < .001$ as ***.}
\label{table 4}
\begin{tabular}{ll|lll|llll|ll}
\toprule
$\#$ ID & Metric &     SHELS &         Cheating &   Ours & \multicolumn{2}{c}{$p_{\mathrm{cheat}}$} & \multicolumn{2}{c|}{$p_{\mathrm{ours}}$} &  $\eta_{\mathrm{cheat}}$ &  $\eta_{\mathrm{ours}}$ \\
classes & & & & &  unadj. & adj. & unadj. & adj. \\
\midrule
\midrule
\multirow{3}{*}{5} & ID &  54.97 {\tiny± 18.84} &     47.22 {\tiny± 11.33} &    30.2 {\tiny± 11.07}**  &            0.30 &                      0.91 &                0.00 &                0.01 &         \multirow{3}{*}{0.62} &               \multirow{3}{*}{0.17} \\
 & OOD &  54.89 {\tiny± 18.26} &     62.22 {\tiny± 21.24} &   86.11 {\tiny± 10.67}**  &            0.44 &                      0.91 &                0.00 &                0.00 & & \\
& Total &   54.93 {\tiny± 6.43} &      54.72 {\tiny± 15.7} &    58.16 {\tiny± 4.72}**  &            0.52 &                      0.91 &                0.01 &                0.01 & &\\
\multirow{3}{*}{6} & ID &   17.94 {\tiny± 1.77} &    10.62 {\tiny± 3.86}** &    10.83 {\tiny± 4.02}**  &            0.00 &                      0.00 &                0.00 &                0.00 &         \multirow{3}{*}{0.13} &               \multirow{3}{*}{0.40} \\
 & OOD &      0.0 {\tiny± 0.0} &  43.68 {\tiny± 20.01}*** &   44.15 {\tiny± 34.39}**  &            0.00 &                      0.00 &                0.00 &                0.01 & & \\
& Total &    9.79 {\tiny± 0.97} &    25.65 {\tiny± 9.17}** &   25.97 {\tiny± 14.36}**  &            0.00 &                      0.00 &                0.00 &                0.01 & &\\
\multirow{3}{*}{7} & ID &    15.51 {\tiny± 2.4} &     9.37 {\tiny± 2.78}** &     8.69 {\tiny± 3.38}**  &            0.00 &                      0.00 &                0.00 &                0.00 &         \multirow{3}{*}{0.11} &               \multirow{3}{*}{0.26} \\
 & OOD &      0.0 {\tiny± 0.0} &  42.52 {\tiny± 17.23}*** &   48.41 {\tiny± 28.14}**  &            0.00 &                      0.00 &                0.00 &                0.00 & & \\
& Total &     9.05 {\tiny± 1.4} &    23.18 {\tiny± 6.68}** &   25.24 {\tiny± 10.26}**  &            0.00 &                      0.00 &                0.00 &                0.01 & &\\
\multirow{3}{*}{8} & ID &   13.96 {\tiny± 3.17} &     7.17 {\tiny± 3.15}** &    6.26 {\tiny± 2.45}***  &            0.00 &                      0.00 &                0.00 &                0.00 &         \multirow{3}{*}{0.10} &               \multirow{3}{*}{0.19} \\
 & OOD &      0.0 {\tiny± 0.0} &  42.56 {\tiny± 17.18}*** &   47.92 {\tiny± 28.94}**  &            0.00 &                      0.00 &                0.00 &                0.00 & & \\
& Total &    8.59 {\tiny± 1.95} &    20.78 {\tiny± 6.42}** &   22.28 {\tiny± 10.23}**  &            0.00 &                      0.00 &                0.00 &                0.01 & &\\
\multirow{3}{*}{9} & ID &    12.0 {\tiny± 2.23} &      7.2 {\tiny± 2.65}** &    5.29 {\tiny± 2.53}***  &            0.00 &                      0.00 &                0.00 &                0.00 &         \multirow{3}{*}{0.30} &               \multirow{3}{*}{0.14} \\
 & OOD &      0.0 {\tiny± 0.0} &  44.94 {\tiny± 16.89}*** &  60.37 {\tiny± 17.21}***  &            0.00 &                      0.00 &                0.00 &                0.00 & & \\
& Total &    7.71 {\tiny± 1.43} &   20.68 {\tiny± 4.99}*** &   24.96 {\tiny± 5.16}***  &            0.00 &                      0.00 &                0.00 &                0.00 & &\\
\bottomrule
\end{tabular}
\end{table}

\end{document}